\documentclass[10pt,letterpaper]{article}
\usepackage[margin=1in]{geometry}
\usepackage{amsmath,amssymb,amsthm}
\usepackage[round]{natbib}

\usepackage{booktabs,tabularx,enumitem}
\usepackage{microtype}
\usepackage[hidelinks]{hyperref}
\usepackage[nameinlink,noabbrev]{cleveref}
\newcommand{\E}{\mathbb{E}}
\newcommand{\R}{\mathbb{R}}
\newcommand{\PP}{\mathcal{P}}

\newcommand{\norm}[1]{\lVert #1\rVert}

\DeclareMathOperator{\Cov}{Cov}

\newtheorem{assumption}{Assumption}
\newtheorem{proposition}{Proposition}
\newtheorem{theorem}{Theorem}

\theoremstyle{remark}

\usepackage{xcolor}

\begin{document}

\title{Neyman-Orthogonal Activation Steering}
\author{David Arbour\\Adobe Research\\\texttt{darbour26@gmail.com} \and Erfan Jahanparast\\MIT\\\texttt{erfanjp@berkeley.edu} \and Serina Chang\\University of California, Berkeley\\\texttt{serinac@berkeley.edu} \and Avi Feller\\University of California, Berkeley\\\texttt{afeller@berkeley.edu}}
\date{}
\maketitle

\begin{abstract}
Activation steering is a popular approach for shifting LLM output toward a desired behavior. The standard recipe estimates a steering direction by contrasting LLM activations on prompts with and without a target concept, such as truthfulness. In this paper, we separate the estimand---the steering direction these methods target---from the estimator used to recover it. We frame common contrastive and probe-based methods as estimators of the average gradient of an outcome regression. This framing clarifies their modeling assumptions and extends steering to continuous outcomes such as log-probabilities or oracle ratings. Flexible ML models that fit the outcome well, however, might not estimate the gradient well. Drawing on the debiased machine learning literature, we develop a complementary estimator built on the Stein score, the Riesz representer of the average derivative. Combining the two routes gives a Neyman-orthogonal estimator with classical double-robustness properties. Researchers can mix and match outcome models and score estimators, including scores from pretrained diffusion models. We illustrate the framework on standard datasets on public opinion, truthfulness, and concept steering, finding that the proposed method improves performance overall.
\end{abstract}
\section{Introduction}
\label{sec:introduction}

Activation steering has emerged as a popular lightweight alternative to fine-tuning for shifting the output of a large language model (LLM) toward a desired behavior. Rather than updating model weights, activation steering methods intervene directly on intermediate activations at inference time, shifting the model toward behaviors such as truthfulness \citep{marks2024geometry}, refusal \citep{arditi2024refusal}, or particular personas \citep{jahanparast2026opinions}. 
The standard approach to activation steering starts by writing prompts that do and do not exhibit a target concept (such as truthfulness) and then recording the activations associated with each prompt. Researchers then estimate a \emph{steering direction} in activation space by learning a relationship between the activations from each prompt, such as by computing the difference in the average activations between on- and off-concept prompts. Finally, researchers add that direction to the generated activations at inference time, typically with the goal of shifting behavior toward the behavior in the on-concept prompt.
Common methods include activation addition~\citep{turner2023activation,rimsky-etal-2024-steering}, inference-time intervention~\citep{li2023inference}, and representation engineering~\citep{zou2023representation}.

In this paper, we show that these approaches share a common if implicit statistical structure in which we can view the steering direction as an estimate of the average gradient of an outcome regression. 
For model activations $X$, with activation marginal $X\sim\PP$ on $\R^d$, and downstream outcome $Y$ (i.e., whether the model exhibits the desired concept), we study the average gradient $\theta_\PP=\E_\PP[\nabla\mu(X)]$ of the outcome regression $\mu(x)=\E[Y\mid X=x]$. 
This is the natural target for first-order steering of a response surface, in which shifting $x$ to $x+\delta v$ changes its average by $\delta v^\top\theta_\PP+O(\delta^2)$. 
Elucidating the underlying regression problem reveals that different steering methods correspond to different choices of outcome model and assumptions about the activation distribution, such as the structure of the covariance matrix. 
Popular approaches like Contrastive Activation Addition (CAA) even replace $Y$ with the binary prompt-set label, $L \in \{0, 1\}$, a workaround which may fail if $L$ is a poor proxy for $Y$, and if $Y$ is continuous.
Approaching steering as a regression problem enables us to immediately extend existing methods to allow for continuous outcomes $Y$ and to leverage flexible machine learning tools when estimating $\mu$.

A model that estimates $\mu$ well, however, may not estimate $\nabla\mu$ well. This motivates estimators that target the average gradient directly. Drawing on average derivative estimation and debiased machine learning~\citep{stoker1986consistent,chernozhukov2018double}, we use the Stein score $\alpha(x)=-\nabla\log p(x)$ of the activation distribution to construct an estimator that does not require differentiating a fitted $\hat\mu$. 
This connection lets us move beyond Gaussian working models by using flexible score estimators, including scores derived from pretrained diffusion models~\citep{luo2026glp}.

We then combine the outcome model and Stein score to obtain the Neyman-orthogonal estimator
\[
\frac{1}{n}\sum_{i=1}^n\left\{\nabla\hat\mu(X_i)
 +\hat\alpha(X_i)\bigl(Y_i-\hat\mu(X_i)\bigr)\right\}.
\]
Researchers can mix and match their choice of $\hat\mu$ and $\hat\alpha$: for example, researchers can pair an average gradient estimated via a linear regression outcome model with an estimated score via kernel density estimation. Importantly, the combined estimator has a product bias, meaning that the estimator is first-order insensitive to errors in either fitted component, which yields classical double robustness under regularity conditions.

All together, our work presents the following contributions:
\begin{itemize}[leftmargin=*]
    \item We unify common activation steering methods under an average-gradient estimation framework. This framework separates the estimand from the estimator, clarifies the assumptions underlying existing methods, and extends steering from binary labels to continuous outcomes.

    \item We draw on debiased machine learning to develop a Stein score-based estimator and a Neyman-orthogonal estimator with double-robustness guarantees. These estimators allow practitioners to combine flexible outcome models with modern score estimators, including scores from pretrained diffusion models.

    \item We evaluate the framework on public opinion and truthfulness steering across six language models, and on concept steering with AxBench. Relative to CAA, Neyman-orthogonal estimators reduce total-variation error by an average of $5.6\%$ on \textsc{OpinionQA} and $16.5\%$ across nine truthfulness benchmarks, with improvements for every model. On AxBench, the overall steering score increases by $8.0\%$, with gains concentrated in text concepts.
\end{itemize}

\subsection{Related work}
\label{sec:related}

\paragraph{Activation steering.}
The most popular form of activation steering is based around contrastive activation addition. Researchers first collect prompts that differ on a binary attribute or concept, take the difference of class means over their activations, and add the resulting direction at generation time~\citep{turner2023activation,rimsky-etal-2024-steering}; see also the literature on the linear representation hypothesis~\citep{park2024linear,park2024geometry}. Related approaches instead intervene at attention heads~\citep{li2023inference} or as part of a broader representation engineering framework~\citep{zou2023representation}. A second line of work refines how directions are constructed, including sparse features~\citep{templeton2024scaling,marks2024sparse}, representation finetuning~\citep{wu2024reft,wu2025reps}, and recursive feature machines~\citep{beaglehole2026universal}. Other methods learn interventions through different objectives, such as by extracting sentence-specific steering vectors \citep{subramani2022extracting} or via distributional matching \citep{rodriguez2025lineas}. See also work on steering reliability and evaluation~\citep{tan2024analysing,wu2025axbench,bas2025whatcan}.

\paragraph{Average derivatives.}
Functionals of the form $\E[\nabla\mu(X)]$ have a long history. In experimental design, \citet{box1951experimental} study this quantity in terms of finding the direction of ``steepest ascent.'' In econometrics, the average derivative is a central target in single-index and semiparametric models~\citep{stoker1986consistent,hardle1989investigating,newey1993efficiency}. Weighted average derivatives are also central in index coefficients~\citep{powell1989semiparametric}. Finally, the score representation is also related to the log-derivative method in reinforcement learning~\citep{williams1992simple}.

\paragraph{Orthogonalization.}
Our main technical machinery comes from automatic debiased machine learning~\citep{chernozhukov2018double,chernozhukov2022econometrica}, which gives a general framework for estimating functionals using their Riesz representers. Closely related average-derivative estimands have been studied within this literature; see, for example \citet{hirshberg2021augmented}. We view this paper as applying these existing frameworks to activation steering. Finally, the term \emph{orthogonality} often has a different connotation in the steering literature, where it is typically used with respect to geometry in concept space \citep{belrose2023leace, park2026information}. These perspectives address different sources of unwanted variation and can be complementary.

\section{Setup and notation}
\label{sec:setup}

We now lay out the steering problem as a statistical estimation problem, separating the \emph{estimand} from the \emph{estimator} used to recover it.

\paragraph{Activations and outcomes.}
Let $W=(X,Y)$ follow a joint law $P$, with activation marginal $X\sim\PP$ on $\R^d$. To simplify exposition, we fix a single layer and token position; for Llama-3.1-8B, for example, $d=4{,}096$. A scalar outcome $Y\in\R$ records the behavior we want the steered model to elicit, such as an oracle rating or the log-probability of a target completion. The \emph{outcome regression} is
\begin{equation}
\mu(x)=\E_P[Y\mid X=x].
\label{eq:mu}
\end{equation}
In the interpretability literature, a fitted regression $\hat\mu$ is known as a \emph{probe}~\citep{alain2017understanding}. 

\paragraph{Estimand: the average gradient.}
The broad goal of \emph{activation steering} \citep{turner2023activation,rimsky-etal-2024-steering,zou2023representation} is to shift $x$ to increase the expected outcome $Y$.
Focusing on local perturbations, we can shift $x$ by $\delta v$ for a unit vector $v$ and small step size $\delta>0$. We can then consider the response-surface objective:
\begin{equation}
J(\delta,v)=\E_\PP[\mu(X+\delta v)].
\label{eq:objective}
\end{equation}
Under smoothness and integrability conditions in Appendix~\ref{app:regularity}, a Taylor expansion gives
\begin{align}
J(\delta,v)-J(0,v)
 &=\delta v^\top\theta_\PP+O(\delta^2),\nonumber\\
\theta_\PP&=\E_\PP[\nabla\mu(X)].
\label{eq:target}
\end{align}
When $\theta_\PP\ne0$, the first-order optimal direction is $v^\star=\theta_\PP/\norm{\theta_\PP}$. In general, we estimate the unnormalized mean gradient and then normalize to recover the direction. 

Importantly,  Equation~\eqref{eq:objective} is defined through the observational regression and need not equal the expected outcome under an activation intervention; see, e.g., \citet{sankaranarayanan2026activation}.
Roughly speaking, Equation~\eqref{eq:objective} will lead to a valid causal change when replacing the selected activations $X$ by $X+\delta v$ changes the expected behavioral outcome according to the same surface $\mu$. 
We believe that a fully causal approach is a promising direction for future work; we retain the simpler object in Equation~\eqref{eq:objective} here because we believe it is useful in clarifying existing approaches. We further evaluate this empirically in Section~\ref{sec:experiments}.

\paragraph{The binary contrastive label.}
The behavioral outcome $Y$ is often hard to measure or only loosely defined. The standard workaround is to proxy it with a prompt-set label $L\in\{0,1\}$. Write $\PP_\ell$ for the distribution of $X\mid L=\ell$ and $\pi=\Pr(L=1)$. For a label-based target, the activation marginal is the mixture
\begin{equation}
\PP=(1-\pi)\PP_0+\pi\PP_1,
\qquad \mu(x)=\Pr(L=1\mid X=x).
\label{eq:mixture}
\end{equation}

By construction, this label $L$ is always observable. Unfortunately, $L$ can often be a poor proxy for $Y$, the behavior of interest. 
For learning a steering direction, $L$ is a useful proxy
for $Y$ when their average gradients point in the
same direction, regardless of their magnitudes.
One sufficient condition for this is the so-called \textit{common single-index model} \citep{powell1989semiparametric}, $\mu^{(Y)}(x)=g_Y(\beta^\top x)$ and $\mu^{(L)}(x)=g_L(\beta^\top x)$, with both links increasing and positive mean derivatives. In this case, both average gradients are positive scalar multiples of $\beta$. The single-index assumption is more fragile when prompts differ along multiple attributes correlated with $L$ but not with $Y$. In that case, the label gradient can track prompt design rather than the behavior of interest.

\section{Outcome and score estimators}
\label{sec:estimators}
\subsection{Outcome models}
\label{sec:outcome-models}

The natural estimator of $\theta_\PP$ first fits a regression $\hat\mu\approx\mu$ on the available $(X,Y)$ data and then averages its gradient over the reference population:
\begin{equation}
\hat\theta_\PP^{\mathrm{plug}}=
\frac{1}{n}\sum_{i=1}^n\nabla\hat\mu(X_i),
\label{eq:plugin}
\end{equation}
with held-out evaluation or cross-fitting when the regression is estimated from the same sample. Table~\ref{tab:first-order-methods} summarizes the resulting directions, which we divide into three groups based on the choice of outcome model: linear regression, logistic regression, and nonlinear models. As we discuss in Section~\ref{sec:orth}, fitting $\hat\mu$ to minimize outcome prediction error need not minimize error in its gradient $\nabla \mu$.

\begin{table*}[t]
\centering
\small
\caption{Regression constructions and their connections to steering methods. $\Delta$ is the difference of class means, $\E[X\mid L=1]-\E[X\mid L=0]$; 
$\Sigma_X=\operatorname{Cov}(X)$ is the marginal activation
covariance; $\Sigma_W$ is the within-class covariance; and
the columns of $U_k$ are the leading eigenvectors of
$\Sigma_X$. Directions are shown up to positive multiplicative constants.}
\label{tab:first-order-methods}
\begin{tabularx}{\textwidth}{@{}lXl@{}}
\toprule
Method or construction & Outcome model or assumption & Direction \\
\midrule
CAA / difference of means & Linear, isotropic working covariance & $\Delta$ \\
OLS & Linear regression of $L$ & $\Sigma_X^{-1}\Delta$ \\
LDA & Gaussian classes, shared covariance & $\Sigma_W^{-1}\Delta$ \\
PCA-projected difference of means & Isotropic working slope, projected onto PCs & $U_kU_k^\top\Delta$ \\
ITI mean-difference direction & Per-head linear, selected attention heads & $\Delta_{(\ell,h)}$ \\
Logistic probe / ITI probe direction & Logistic, global or per-head & $\hat\beta^{\mathrm{logit}}$ \\
Kernel mean-gradient estimator & Differentiable nonlinear regression & $\E_\PP[\nabla\hat\mu(X)]$ \\
\bottomrule
\end{tabularx}
\end{table*}

\paragraph{Linear regression-based methods.}
Several common steering directions can be represented through linear working models for $L$ given $X$.
For ordinary least squares, $\hat\mu(x)=\hat\beta^\top x+\hat\beta_0$, the gradient is simply the coefficient vector $\hat\beta$. With a binary outcome $L$ and invertible pooled covariance, the population slope satisfies
\begin{align}
\beta_{\mathrm{OLS}}&=\pi(1-\pi)\Sigma_X^{-1}\Delta,\nonumber\\
\Delta&=\E[X\mid L=1]-\E[X\mid L=0].
\label{eq:ols}
\end{align}
If we restrict the working covariance $\Sigma_X$ to be isotropic, we obtain the difference-of-means direction underlying the popular contrastive activation addition~\citep{rimsky-etal-2024-steering,turner2023activation}.
Penalized regression and covariance regularization offer other ways to estimate the direction when the number of activations is small relative to their dimension. Section~\ref{sec:stein-score} suggests an alternative route to this equality under Gaussian activations.

\citet{marks2024geometry} give a related framing via Fisher's linear discriminant analysis~\citep{fisher1936use}. Under shared-covariance Gaussian classes, the learned direction is $\Sigma_W^{-1}\Delta$. Since $\Sigma_X=\Sigma_W+\pi(1-\pi)\Delta\Delta^\top$, the OLS and LDA directions are proportional; see Appendix~\ref{app:worked-example}. PCA-projected difference of means, $U_kU_k^\top\Delta$, is another regularized direction; unlike principal component regression, this approach does not inverse-weight the retained eigenvalues. Finally, RepE's LAT baseline instead uses the first principal component of paired activation differences as a reading vector~\citep{zou2023representation}.

\paragraph{Logistic regression-based methods.}
An alternative is to fit $\hat\mu(x)=\text{logit}^{-1}(\hat\beta^\top x+\hat\beta_0)$~\citep{alain2017understanding}. Since
$\nabla\hat\mu(x)=\hat\mu(x)\{1-\hat\mu(x)\}\hat\beta$,
the mean gradient is a nonnegative scalar times the logistic coefficient, and has the same direction when this scalar is positive. Under the shared-covariance Gaussian class model, correctly specified logistic regression and LDA target the same direction in the population limit, though they differ in finite samples~\citep{efron1975efficiency}. Finally, ITI selects attention heads by validation probe accuracy and intervenes using probe-weight or class-mean-difference directions~\citep{li2023inference}.

\paragraph{Nonlinear outcome models.}
Finally, we can estimate nonlinear outcome models, such as kernel ridge regression. Recursive feature machines (RFMs) fit kernel regressions while updating a learned metric using gradient outer products~\citep{radhakrishnan2024mechanism,beaglehole2026universal}. For such outcome models, the gradient $\nabla\hat\mu(x)$ varies with $x$, so estimating a single global direction requires averaging over $\PP$. Fortunately, we can still apply Equation~\eqref{eq:plugin}, which computes the average gradient for any differentiable fitted regression, including an RFM. In Appendix \ref{sec:agop}, we also consider learning subspaces rather than directions, focusing on the average gradient outer product (AGOP), as in \citet{beaglehole2026universal}.

\subsection{Stein score-based methods}
\label{sec:stein-score}

An alternative route to estimating $\theta_\PP$ uses the activation distribution rather than the derivative of a fitted outcome model. Stein's lemma, in its simplest form, shows that $\E[Xg(X)]=\E[g'(X)]$ for $X\sim N(0,1)$ and suitable smooth functions $g$~\citep{stein1972bound}. The Stein score generalizes this idea. Suppose $\PP$ has a differentiable density $p$, and define $\alpha(x)=-\nabla\log p(x)$. Integration by parts gives
\begin{equation}
\E_\PP[\nabla g(X)]=\E_\PP[\alpha(X)g(X)]
\label{eq:stein-identity}
\end{equation}
for functions with integrable derivatives and vanishing boundary terms; see Appendix~\ref{app:regularity}.

Applying this result to the outcome regression yields
\begin{equation}
\theta_\PP=\E_\PP[\alpha(X)\mu(X)]
            =\E_P[\alpha(X)Y].
\label{eq:riesz-pop}
\end{equation}
Thus, $\alpha$ is the coordinatewise Riesz representer of the average derivative, connecting steering to methods for estimating Riesz representers~\citep{chernozhukov2022econometrica}. Plugging in the estimated score gives
\begin{equation}
\hat\theta_\PP^{\mathrm{Stein}}
 =\frac{1}{n}\sum_{i=1}^n\hat\alpha(X_i)Y_i.
\label{eq:score-est}
\end{equation}

As in the broader causal machine learning literature, the outcome and weighting perspectives are complementary. If $X\sim N(m,\Sigma)$, then $\alpha(x)=\Sigma^{-1}(x-m)$ and $\theta_\PP=\Sigma^{-1}\Cov(X,Y)$ for any sufficiently smooth $\mu$. The population least-squares slope therefore recovers the average gradient under either a linear true regression or a Gaussian activation marginal. For a binary label, an isotropic working covariance gives the CAA direction. For Gaussian class distributions, however, this generally yields a non-Gaussian mixture; see Appendix~\ref{app:worked-example}.

\paragraph{Estimating the score.}
We can estimate $\alpha$ by a range of methods, including density estimation, score matching~\citep{hyvarinen2005}, and direct Riesz regression. Under the conditions in Appendix \ref{app:regularity}, the population score-matching objective is
\begin{equation}
\mathcal L(a)=\E_\PP\left[\norm{a(X)}^2
                      -2\nabla\!\cdot a(X)\right].
\label{eq:score-loss}
\end{equation}
Applying the Stein identity coordinatewise gives
\[
\mathcal L(a)
=
\E_\PP\!\left[\|a(X)-\alpha(X)\|^2\right]
-
\E_\PP\!\left[\|\alpha(X)\|^2\right].
\]
Thus the minimizer is $\alpha$ whenever it belongs to
the candidate class.
Unlike fitting an outcome model, estimating $\hat\alpha$ does not require outcome labels. Instead, the same activation-based score model can apply to any number of outcomes on the same population, analogous to how inverse propensity score weights can be re-applied to multiple outcomes.

In practice, flexible score estimation is challenging in high-dimensional activation spaces, making regularization important. One promising avenue is pretrained models, which offer a practical means of leveraging larger activation datasets. For example, the Generative Latent Prior (GLP) of \citet{luo2026glp} trains a diffusion model on roughly a billion residual-stream activations without using any steering-task labels. We can leverage this GLP by calibrating the pretrained score to each particular dataset.

\section{Orthogonalized learning}
\label{sec:orth}

Following the literature on automatic debiased machine learning~\citep{chernozhukov2022econometrica}, we can combine the outcome model and Stein score approaches to obtain an estimator with less sensitivity to errors in either separately. Recognizing the target as an average derivative, we directly apply existing AutoDML constructions~\citep{hirshberg2021augmented}, starting with the fitted gradient and adding a score-weighted residual:
\begin{align}
\hat\theta_\PP^{\mathrm{orth}}
 =\frac{1}{n}\sum_{i=1}^n\bigl[&\nabla\hat\mu_{-k(i)}(X_i)\nonumber\\
 &+\hat\alpha_{-k(i)}(X_i)
       \{Y_i-\hat\mu_{-k(i)}(X_i)\}\bigr].
\label{eq:orth}
\end{align}
Here $\hat\mu_{-k(i)}$ and $\hat\alpha_{-k(i)}$ are cross-fit without the fold containing observation $i$. 
Setting $\hat\alpha\equiv0$ recovers the plug-in estimator; setting $\hat\mu\equiv0$ recovers the score estimator.

\paragraph{Product bias.}
To understand this orthogonalization property, suppose first that the score is exact. Holding the fitted regression $m$ fixed, Equation~\eqref{eq:stein-identity} gives
\begin{equation}
\E_P[\alpha(X)\{Y-m(X)\}]
=\theta_\PP-\E_\PP[\nabla m(X)].
\label{eq:correction-mean}
\end{equation}
The expected residual correction is exactly the difference between the target and the mean fitted gradient.
If the regression is exact instead, the residual has conditional mean zero, so adding the correction leaves the target unchanged.

More generally, let $\tilde{\theta}_\PP(m,a)=\E_P[\nabla m(X)+a(X)\{Y-m(X)\}]$ for fixed fitted functions $m$ and $a$. Applying the Stein identity to $m-\mu$ yields
\begin{equation}
\tilde{\theta}_\PP(m,a)-\theta_\PP
=-\E_\PP[(a-\alpha)(m-\mu)].
\label{eq:bias}
\end{equation}
Errors in one nuisance enter the bias only multiplied by errors in the other. This product structure gives Neyman orthogonality and classical double robustness, in the sense that the population bias vanishes if either nuisance error converges to zero, while the other remains bounded in $L^2(\PP)$. See Appendix~\ref{app:regularity}.

\paragraph{A Gaussian example.}
To fix ideas, let $X\sim N(0,\Sigma)$, with mean-gradient target $\theta_\PP$. Consider a fitted linear regression $m(x)=b^\top x+c$ and a working score $a(x)=Ax$. Stein's identity gives $\Cov(X,Y)=\Sigma\theta_\PP$, so
\begin{align}
\tilde{\theta}_\PP(m,a)&=b+A\Cov(X,Y)-A\Sigma b\nonumber\\
      &=\theta_\PP+(I-A\Sigma)(b-\theta_\PP).
\label{eq:gaussian-main}
\end{align}
An exact score, $A=\Sigma^{-1}$, removes the slope error for any $b$; a correct slope removes the bias for any $A$. For example, ridge regression generally shrinks different components of the slope by different amounts, changing its direction. 
With a correctly-specified score, the correction removes this population
distortion; see Appendix~\ref{app:worked-example}.

\paragraph{Practical construction.} 
The proposed framework allows researchers to mix and match a large set of outcome and score estimators. 
In practice, we suggest starting from a robust outcome model, such as CAA fit as a calibrated linear regression, and treating the score as a correction to that model.
We also recommend fitting both nuisances outside each evaluation fold, keeping paired prompts together. We then estimate the score in a low-dimensional projection of the activations, where density estimation is tractable.
Finally, comparing the plug-in and corrected directions
provides a diagnostic of sensitivity to the score correction. We further assess its effect on downstream behavior using held-out
validation data.

\paragraph{Orthogonalized AGOP.}
Finally, we can apply the same orthogonalization principle to the average gradient outer product, which summarizes variation across gradient directions \citep{beaglehole2026universal}. Unlike with learning the average gradient, the correction term for the AGOP is matrix-valued and includes a Hessian term. Appendix~\ref{sec:agop} develops the Neyman-orthogonal AGOP estimator.

\section{Experiments}
\label{sec:experiments}

The overall goal of our experiments is to assess whether the Neyman estimator in Equation~\eqref{eq:orth} improves performance relative to outcome modeling alone. We also demonstrate how our framework enables researchers to learn steering directions with continuous responses $Y$ in place of binary contrastive labels $L$.

Throughout the experiments, we compare the plug-in estimator~\eqref{eq:plugin} with the Neyman estimator~\eqref{eq:orth} built on the same outcome model. 
Our primary outcome model is a calibrated linear predictor
with slope in the CAA difference-of-means direction.
We also evaluate linear ridge regression, kernel ridge
regression, and, for OpinionQA, recursive feature machines.
For the score model, we consider: an isotropic Gaussian score; a kernel density estimator (KDE) with a Gaussian kernel; and a random Fourier feature (RFF) approximation. 
We use held-out validation data to select the steering
configuration, with benchmark-specific choices described
in Appendix~\ref{app:experiments}.
We also use a held-out validation set to select the layer for OpinionQA and truthfulness; for AxBench, we fix the layer at ten following the original benchmark protocol. We evaluate all estimators on the same splits and test items, and we report paired bootstrap intervals for the aggregate contrasts identified in the table captions. See Appendix~\ref{app:experiments} for the full protocol.

\paragraph{Public opinion steering}
\label{sec:exp-opinion}
Our first experiment takes a public opinion dataset, \textit{OpinionQA}~\citep{santurkar2023whose}, with the task of steering a model toward the opinion distribution of Republican or Democratic respondents.
We ran on Qwen3-8B~\citep{yang2025qwen3} and Qwen3.5-9B~\citep{qwen2026qwen35}, each in base and instruct releases, together with Qwen3-4B-Base and Llama-3.1-8B~\citep{llama3}, to measure across model size, model family, and post-training.
We constructed the outcome by taking the mean response for each group (i.e., Republicans, Democrats). 
The quality of the steering was measured by taking the total variation and KL divergence between the steered model's distribution over answer options and the target group's observed distribution.
Additional experiment configuration details are provided in Appendix \ref{app:experiments}.

\begin{table*}[t]
  \centering
  \small
  \setlength{\tabcolsep}{4.5pt}
  \caption{OpinionQA partisan steering, test error averaged over five seeds; lower is better. ``Unsteered'' is the model on the neutral question. ``Gain'' is the relative reduction from CAA to the Neyman estimator in each setting. The aggregate gain is the unweighted mean of the twelve setting-specific relative reductions;
  the 95\% interval resamples the five seeds, which share splits across settings.}
  \label{tab:opinion}
  \begin{tabular}{llccccccrr}
    \toprule
    & & \multicolumn{3}{c}{TV $\downarrow$} & \multicolumn{3}{c}{KL $\downarrow$} & \multicolumn{2}{c}{Gain (\%)} \\
    \cmidrule(lr){3-5}\cmidrule(lr){6-8}\cmidrule(lr){9-10}
    Model & Target & Unsteered & CAA & Neyman & Unsteered & CAA & Neyman & TV & KL \\
    \midrule
    Llama-3.1-8B    & Dem & 0.271 & 0.265 & \textbf{0.263} & 0.254 & 0.229 & \textbf{0.216} & 0.6 & 5.7 \\
                    & Rep & 0.301 & 0.276 & \textbf{0.259} & 0.303 & 0.237 & \textbf{0.218} & 6.0 & 8.0 \\
    Qwen3-4B-Base   & Dem & 0.241 & 0.232 & \textbf{0.231} & 0.209 & 0.192 & \textbf{0.176} & 0.6 & 8.1 \\
                    & Rep & 0.315 & 0.270 & \textbf{0.269} & 0.356 & 0.266 & \textbf{0.257} & 0.3 & 3.5 \\
    Qwen3-8B        & Dem & 0.354 & 0.308 & \textbf{0.284} & 0.725 & 0.442 & \textbf{0.323} & 7.8 & 26.9 \\
                    & Rep & 0.431 & 0.381 & \textbf{0.266} & 1.067 & 0.514 & \textbf{0.241} & 30.3 & 53.2 \\
    Qwen3-8B-Base   & Dem & 0.227 & 0.219 & \textbf{0.214} & 0.185 & 0.175 & \textbf{0.158} & 2.3 & 9.8 \\
                    & Rep & 0.303 & 0.284 & \textbf{0.268} & 0.343 & 0.274 & \textbf{0.231} & 5.5 & 15.9 \\
    Qwen3.5-9B      & Dem & 0.316 & 0.278 & \textbf{0.267} & 0.417 & 0.286 & \textbf{0.256} & 4.2 & 10.5 \\
                    & Rep & 0.380 & 0.303 & \textbf{0.294} & 0.594 & 0.352 & \textbf{0.272} & 3.1 & 22.7 \\
    Qwen3.5-9B-Base & Dem & 0.302 & 0.257 & \textbf{0.252} & 0.378 & 0.245 & \textbf{0.233} & 1.7 & 5.2 \\
                    & Rep & 0.373 & 0.292 & \textbf{0.279} & 0.597 & 0.268 & \textbf{0.256} & 4.3 & 4.6 \\
    \midrule
    Average         &     & 0.318 & 0.280 & \textbf{0.262} & 0.452 & 0.290 & \textbf{0.236} & 5.6 {\scriptsize[4.8, 6.3]} & 14.5 {\scriptsize[13.5, 15.5]} \\
    \bottomrule
  \end{tabular}
\end{table*}

Table~\ref{tab:opinion} shows the results.
We see that the Neyman estimator lowers average test error relative to CAA across all six models, two targets, and two metrics.
The mean reduction is $5.6\%$ for TV (95\% CI $[4.8, 6.3]$) and $14.5\%$ for KL ($[13.5, 15.5]$). We see the largest gains when the outcome-only model is weakest; for example, steering Qwen3-8B toward Republican opinions, TV falls from $0.381$ to $0.266$, and KL falls from $0.514$ to $0.241$.
We still see some modest improvement when outcome-only model steering is already strong, as on Llama-3.1-8B and Qwen3-4B-Base toward Democratic opinions.
The same pattern holds when the outcome model is linear ridge, KRR, or RFM rather than CAA (Appendix Table~\ref{tab:opinion-outcomes}). For each, the Neyman estimator lowers average error relative to its own plug-in on every model, by $6.2$--$6.4\%$ in TV and $14.3$--$15.5\%$ in KL, with the largest gains again on Qwen3-8B.

\paragraph{Truthfulness steering}
\label{sec:exp-truth}
Our second set of experiments was run on the ``truthfulness'' task, in which the steering task is to move the model's answers on true/false statements toward the correct label.
We use nine datasets considered in prior work~\citep{marks2024geometry}: \textsc{Cities} and \textsc{Sp-En Trans.}, their negated variants \textsc{Neg Cities} and \textsc{Neg Sp-En Trans.}, the compositional \textsc{Cities Conj.}\ and \textsc{Cities Disj.}, and \textsc{Companies}, \textsc{Common Claim}, and \textsc{Counterfact}.
We ran on the same six models as in OpinionQA, with two seeds per model and a 60/20/20 split of each dataset.
Each training statement $s$ appears under three prompt variants $t$: neutral, instructed to answer truthfully, and instructed to lie.
During training we also subtracted the mean of all three prompt variants per statement from the activations to control for erroneous statement level variation.
For the outcome $Y$, we constructed a continuous metric by taking the model's margin toward the correct answer under the same statement variant, i.e., 
\begin{equation}
  Y_{st} \;=\; (2\ell_s-1)\,\bigl\{z_{\textsc{true}}\bigl(q_t(s)\bigr)-z_{\textsc{false}}\bigl(q_t(s)\bigr)\bigr\},
  \label{eq:witness}
\end{equation}
where $q_t(s)$ is the variant-$t$ prompt ending in ``This statement is:'', $\ell_s\in\{0,1\}$ is the truth label, and $z_{\textsc{true}}$ is the largest next-token logit among the tokens for \textsc{true}, and likewise for \textsc{false}.
Following prior work, we report TV and KL to the correct answer, taking the model's probability of the correct answer to be the logistic transform of the true/false margin on the neutral test prompt.

\begin{table}[t]
  \centering
  \small
  \setlength{\tabcolsep}{2.6pt}
  \caption{Truthfulness with CAA as the outcome model: test TV averaged over nine datasets and two seeds, then the all-model averages of TV and KL; lower is better. Q: Qwen; B: base; ``Unst.'': unsteered. ``Probe'' is the label mass-mean direction; ``Stein'' is the score-only estimator with the isotropic score on demeaned activations. ``Gain'' is the relative reduction in mean error from CAA to the Neyman estimator, weighting dataset--seed units equally. All-model 95\% intervals, resampling the nine datasets, are $[11, 24]$ for TV and $[17, 30]$ for KL.}
  \label{tab:truth}
  \begin{tabular}{lccccc r}
    \toprule
    Model & Unst. & Probe & Stein & CAA & Ney. & Gain \\
    \midrule
    Llama-3.1-8B & .492 & .463 & .465 & .459 & \textbf{.444} & 3.2 \\
    Q3-4B-B & .495 & .423 & .334 & .323 & \textbf{.281} & 12.8 \\
    Q3-8B & .475 & .394 & .225 & .298 & \textbf{.195} & 34.6 \\
    Q3-8B-B & .452 & .424 & .350 & .393 & \textbf{.283} & 28.0 \\
    Q3.5-9B & .432 & .367 & .312 & .294 & \textbf{.271} & 7.9 \\
    Q3.5-9B-B & .461 & .415 & .336 & .372 & \textbf{.312} & 16.0 \\
    \midrule
    All TV & .468 & .414 & .337 & .356 & \textbf{.298} & 16.5  \\
    All KL & 1.202 & .619 & .485 & .541 & \textbf{.419} & 22.5  \\
    \bottomrule
  \end{tabular}
\end{table}

\begin{table}[t]
  \centering
  \small
  \setlength{\tabcolsep}{2.6pt}
  \caption{Truthfulness with linear ridge and KRR outcome models fit to $Y$: plug-in (``Plug'') and Neyman (``Ney.'') estimators built on each, same averaging as Table~\ref{tab:truth}; lower is better. ``Gain'' is the relative reduction in mean error from the plug-in to the Neyman estimator. All-model 95\% intervals, resampling the nine datasets: linear $[15, 34]$ (TV) and $[19, 42]$ (KL); KRR $[16, 34]$ and $[17, 40]$.}
  \label{tab:truth-outcomes}
  \begin{tabular}{lccr|ccr}
    \toprule
    & \multicolumn{3}{c|}{Linear} & \multicolumn{3}{c}{KRR} \\
    \cmidrule(lr){2-4}\cmidrule(lr){5-7}
    Model & Plug & Ney. & Gain & Plug & Ney. & Gain \\
    \midrule
    Llama-3.1-8B & .471 & \textbf{.448} & 4.9  & .469 & \textbf{.444} & 5.2 \\
    Q3-4B-B & .430 & \textbf{.322} & 25.1 & .432 & \textbf{.326} & 24.6 \\
    Q3-8B & .367 & \textbf{.213} & 42.0 & .383 & \textbf{.208} & 45.8 \\
    Q3-8B-B & .404 & \textbf{.324} & 19.8 & .390 & \textbf{.313} & 19.5 \\
    Q3.5-9B & .378 & \textbf{.288} & 23.8 & .366 & \textbf{.286} & 21.7 \\
    Q3.5-9B-B & .421 & \textbf{.313} & 25.7 & .415 & \textbf{.308} & 25.8 \\
    \midrule
    All TV & .412 & \textbf{.318} & 22.8  & .409 & \textbf{.314} & 23.2  \\
    All KL & .630 & \textbf{.443} & 29.8  & .607 & \textbf{.437} & 28.0  \\
    \bottomrule
  \end{tabular}
\end{table}

Tables~\ref{tab:truth} and~\ref{tab:truth-outcomes} show the results.
Here we see that the Neyman estimator lowers average error relative to its plug-in counterpart for every outcome model and every language model.
Averaged over models, TV falls by $16.5\%$ for CAA (95\% CI $[11, 24]$), $22.8\%$ for linear ridge ($[15, 34]$), and $23.2\%$ for KRR ($[16, 34]$).
As before, we see the largest gains when the outcome-only model is weakest. In particular, linear and KRR start well behind CAA, but all three land within $0.02$ in TV after correction. 
The Neyman estimator built on CAA is the best estimator overall, with mean TV falling from $0.356$ to $0.298$ and mean KL from $0.541$ to $0.419$.
Mean TV also falls on each of the nine datasets, from $6.0\%$ on \textsc{Common Claim} to $45.5\%$ on \textsc{Cities} (Table~\ref{tab:truth-datasets}).

\paragraph{AxBench}
\label{sec:exp-axbench}
Our final experiment uses AxBench~\citep{wu2025axbench}.
Following the original paper, we use the Concept500 data with Gemma-2-2B-it~\citep{gemmateam2024gemma2} and intervene on the 10th layer's residual stream with unit-norm directions.
We report the overall steering score proposed by \citet{wu2025axbench}.
More specifically, the score is the harmonic mean of LLM judged concept, instruction, and fluency scores, each ranging between 0 and 2.
We note that we used Qwen3-8B as the judge in place of GPT-4o-mini due to compute limitations, so our scores are comparable with each other but not with the public AxBench leaderboard.
Finally, because the Concept500 dataset consists of prompts from three substantively different question genres (text, code, or math), we measured both the overall steering scores as well as per-genre scores.
Additional experiment configuration details may be found in Appendix \ref{app:experiments}.

\begin{table}[t]
  \centering
  \small
  \setlength{\tabcolsep}{4pt}
  \caption{AxBench Concept500, Gemma-2-2B-it, layer 10. Overall steering (0--2; higher is better) at the validation-selected strength, by concept genre (number of concepts in parentheses). ``Prompt'' instructs the model to include the concept, with no intervention; for code and math, validation selects CAA, so the two columns coincide by construction. The ``All'' row averages concepts equally. Intervals are 95\% paired bootstrap intervals over concepts.}
  \label{tab:axbench}
  \begin{tabular}{lcccc}
    \toprule
    Genre & Prompt & CAA & Neyman & Gain \\
    \midrule
    Text (332) & 0.523 & 0.665 & \textbf{0.740} & $+.075$ {\scriptsize[.044,.106]} \\
    Code (122) & 0.335 & 0.559 & 0.559 & --- \\
    Math (46)  & 0.409 & 0.499 & 0.499 & --- \\
    \midrule
    All (500)  & 0.466 & 0.624 & \textbf{0.673} & $+.050$ {\scriptsize[.029,.070]} \\
    \bottomrule
  \end{tabular}
\end{table}

Table~\ref{tab:axbench} shows the results.
Across all 500 concepts the composite steering score rises from $0.624$ to $0.673$ ($+8.0\%$).
The gain comes entirely from text concepts, where the score rises from $0.665$ to $0.740$. 
For code and math, the validation approach selects CAA, so the two columns coincide by construction.

\section{Discussion}
\label{sec:discussion}

We revisit the growing literature on activation steering from the perspective of causal machine learning. We argue that many existing methods for learning the steering direction correspond to estimating the average gradient of an outcome regression under particular choices of outcome and functional form. 
A key consequence is that such methods implicitly steer toward the presence of a concept in a prompt, which might not align with the downstream behavior itself. 

After clarifying the estimand, we then leverage causal machine learning methods to estimate the average gradient with flexible ML tools.
We first estimate the average gradient using the Stein score,
the Riesz representer of the average-derivative functional.
We then propose a Neyman-orthogonal estimator that combines both outcome modeling and the Stein score, yielding a doubly robust estimator with reduced sensitivity to errors in the nuisance functions.

Finally, we assess the proposed methods on several activation steering benchmarks, finding improved performance overall. 
We find particularly large gains in truthfulness steering, with lower average error for every language model and outcome model considered.

\paragraph{Limitations.} While activation steering is a promising direction for modifying model behavior, the performance of existing approaches, including ours, remains very sensitive to implementation details. For example, even small changes in setting the grid for hyperparameter tuning can lead to non-trivial changes in performance. An especially challenging implementation choice is selecting the appropriate layer and set of activations to steer more broadly. More broadly, selecting a single direction in high-dimensional space is an inherently demanding statistical task. And while our proposed dimension-reduction strategies are promising, they are yet more implementation decisions to make. Finally, we have largely taken measurement of downstream outcomes $Y$ as given. In practice, how best to measure model behavior is an open question, with meaningful limits to the LLM-as-judge paradigm.

\paragraph{Future directions.} There are many promising avenues for future work. The most immediate is that we have focused on a single, global steering direction applied uniformly across prompts. A natural refinement is to instead consider conditional average derivatives, such as $\theta(z)=\E[\nabla\mu(X)\mid Z=z]$ for (groups of) prompts, such as prompt type~\citep{hsu2026contextual,davarmanesh2026attention}. This is a much more demanding statistical task, analogous to estimating the Conditional Average Treatment Effect versus the overall ATE, and it is not clear how to directly extend our proposed methods to this more challenging setting.

Another promising direction is to view prompting for activation steering through the lens of experimental design in high dimensions. Right now, prompt choice for activation steering is often \emph{ad hoc}, with researchers specifying pairs of on- and off-concept prompts. With the additional flexibility of our proposed framework, researchers can instead consider complex, possibly factorial prompt designs, which could improve the performance of learning robust steering directions.

\section*{AI Use Statement}
AI tools assisted with editing, writing of experiment code, suggesting and checking of derivations, and preparing the \LaTeX{} source.
\bibliographystyle{abbrvnat}
\bibliography{steering_refs}
\clearpage
\appendix
\section{Formal results}
\label{app:regularity}

We give conditions for the orthogonalized estimator in Section~\ref{sec:orth}. Throughout, $W=(X,Y)$ has joint distribution $P$, and $X$ has marginal $\PP$. Let $\mu$ be a specified differentiable version of the outcome regression $\E_P[Y\mid X=x]$ and write $\|f\|_2^2=\E_\PP[\|f(X)\|^2]$.

\begin{assumption}[Average-derivative representation]
\label{ass:app-stein}
Assume $\E_P[Y^2]<\infty$. There is a fixed linear class $\mathcal H$ of functions differentiable on a neighborhood of the support of $\PP$, containing constants, $\mu$, and the fitted regressions. Every $f\in\mathcal H$ satisfies $\|f\|_2<\infty$ and $\E_\PP[\|\nabla f(X)\|]<\infty$. There exists $\alpha\in L^2(\PP)^d$ such that, for every $f\in\mathcal H$,
\begin{equation}
\E_\PP[\nabla f(X)]=\E_\PP[\alpha(X)f(X)].
\label{eq:app-representation}
\end{equation}
\end{assumption}

The assumption jointly restricts the activation distribution and the outcome-model class. We retain the notation $\alpha$ for a representer satisfying~\eqref{eq:app-representation}, although it does not need to be a density score. In particular, the assumption implies that $\E_\PP[\nabla h(X)]=0$ whenever $h\in\mathcal H$ vanishes $\PP$-almost surely. 
Applying the identity to $\mu$ and taking conditional expectations also gives $\theta_\PP=\E_P[\alpha(X)Y]$, so the score-only representation extends to this setting. 

The formal results below directly use this assumption, and we next note several sufficient conditions to satisfy these restrictions.
In the main text, we separately conduct extensive  experiments to assess the methods' practical performance in settings where this assumption may not hold.

\begin{itemize}
\item \textbf{Smooth activation densities.}
The Stein score gives one sufficient construction. If $\PP$ has a differentiable density $p$, $\alpha=-\nabla\log p\in L^2(\PP)^d$, and integration by parts has vanishing boundary terms for every $f\in\mathcal H$, then~\eqref{eq:app-representation} holds. For example, a nonsingular Gaussian law satisfies this identity for smooth functions with sufficiently slow growth.

\item \textbf{Linear outcome models.} Alternatively, we can restrict the outcome model without requiring an activation density.
Let $\mathcal H=\{x\mapsto c+b^\top x\}$, and suppose $X$ has finite second moments and nonsingular covariance $\Sigma$. With $\bar x=\E_\PP[X]$, take
\begin{equation}
\alpha(x)=\Sigma^{-1}(x-\bar x).
\label{eq:app-linear-representer}
\end{equation}
For $f(x)=c+b^\top x$, we have $\E_\PP[\nabla f(X)]=b$ and
$\E_\PP[\alpha(X)f(X)]=\Sigma^{-1}\Sigma b=b$.
This calculation holds for discrete as well as continuous activation laws. If the true regression and fitted regressions instead belong to
$\mathcal H=\{x\mapsto c+b^\top B^\top x\}$ for a fixed
$d\times r$ matrix $B$, it is enough that
$\Sigma_B=\operatorname{Cov}_{\PP}(B^\top X)$ be nonsingular.
A valid representer is then
\[
\alpha_B(x)=B\Sigma_B^{-1}B^\top(x-\bar x).
\]

\item \textbf{Fixed nonlinear features.}
A broader construction allows $\mathcal H=\{x\mapsto b^\top\phi(x):b\in\R^p\}$ for a fixed differentiable feature map $\phi$, with constants in its span. Define the $d\times p$ Jacobian $J_\phi(x)=[\nabla\phi_1(x),\ldots,\nabla\phi_p(x)]$ and
\[
G=\E_\PP[\phi(X)\phi(X)^\top],
\qquad D=\E_\PP[J_\phi(X)].
\]
If $\E_\PP\|\phi(X)\|^2<\infty$, $\E_\PP\|J_\phi(X)\|_F<\infty$, and $G$ is nonsingular, then
\begin{equation}
\alpha(x)=DG^{-1}\phi(x)
\label{eq:app-feature-representer}
\end{equation}
satisfies~\eqref{eq:app-representation}.
This includes fixed-degree polynomial models, differentiable splines with fixed knots, and neural features with a fitted linear output layer. For singular $G$, the construction $\alpha=DG^\dagger\phi$
remains valid provided $\ker G\subseteq\ker D$.

\end{itemize}

\paragraph{Exact bias.}
For a regression $m$ and vector-valued working representer $a$, define
\begin{equation}
q(W;m,a)=\nabla m(X)+a(X)\{Y-m(X)\}.
\label{eq:app-moment}
\end{equation}
For fitted $m$ and $a$, the expectations below condition on their training sample and average over independent observations.

\begin{proposition}[Product bias]
\label{prop:exact-bias}
Under Assumption~\ref{ass:app-stein}, for $m\in\mathcal H$ and $a\in L^2(\PP)^d$ with the required integrability,
\begin{equation}
\E_P[q(W;m,a)]-\theta_\PP
=-\E_\PP[(a-\alpha)(m-\mu)],
\qquad
\bigl\|\E_P[q(W;m,a)]-\theta_\PP\bigr\|
\leq \|a-\alpha\|_2\|m-\mu\|_2.
\label{eq:app-bias}
\end{equation}
\end{proposition}

\begin{proof}
Let $h=m-\mu\in\mathcal H$. Iterated expectation gives $\E_P[a(X)\{Y-m(X)\}]=-\E_\PP[ah]$. Subtracting $\theta_\PP=\E_\PP[\nabla\mu]$ from the first term of~\eqref{eq:app-moment}, and applying~\eqref{eq:app-representation} to $h$, gives
\[
\E_P[q(W;m,a)]-\theta_\PP
=\E_\PP[\nabla h-ah]
=\E_\PP[(\alpha-a)h].
\]
Finally, we can apply Cauchy--Schwarz to obtain the bound.
\end{proof}

The model-double robustness property is that the population bias is zero if either nuisance is correct. The Neyman-orthogonality property is that, along admissible perturbations $m=\mu+t h$ with $h\in\mathcal H$ and $a=\alpha+t u$, the bias equals $-t^2\E_\PP[uh]$, with zero derivative at $t=0$.

\paragraph{An externally fitted representer.}
The same identity applies to an activation-based weight $a$ fitted on external data independently of the evaluation sample, including a working score from a pretrained model. Conditional on the fitted functions, let $r_a=\|a-\alpha\|_2$, measuring error under the target marginal $\PP$. Equation~\eqref{eq:app-bias} bounds the population bias by $r_a\|m-\mu\|_2$. Thus an $L^2(\PP)$-consistent regression in $\mathcal H$ removes this bias when $r_a=O_P(1)$, even if the external training distribution differs from $\PP$.

\paragraph{Cross-fitting and inference.}
Let $W_1,\ldots,W_n$ be independent draws from $P$. 
Partition the indices independently of the observations
into a fixed number $K$ of folds $I_k$, with $n_k=|I_k|$
and $n_k/n\to\rho_k>0$. Let $k(i)$ denote the fold
containing observation $i$. Fit $m_k,a_k$ without using
fold $k$, and evaluate $q_k(W)=q(W;m_k,a_k)$ on that fold.
The estimator is $\hat\theta=n^{-1}\sum_i q_{k(i)}(W_i)$. Write $q_0(W)=q(W;\mu,\alpha)$ and define
\begin{equation}
\psi_0(W)=q_0(W)-\theta_\PP
=\nabla\mu(X)+\alpha(X)\{Y-\mu(X)\}-\theta_\PP.
\label{eq:eif}
\end{equation}

We can then make standard assumptions on moment stability and product rate.

\begin{assumption}[Moment stability and product rate]
\label{ass:crossfit}
Dimension $d$ is fixed, $\E_P\|q_0(W)\|^2<\infty$, and, for each fold,
\[
\|q_k-q_0\|_{L^2(P)}=o_P(1),
\qquad
\|a_k-\alpha\|_2\|m_k-\mu\|_2=o_P(n^{-1/2}).
\]
The $L^2(P)$ norm integrates over a fresh observation while holding the fitted functions fixed.
\end{assumption}

Moment stability further requires control of derivatives as well as prediction error. Writing $h_k=m_k-\mu$ and $e_k=a_k-\alpha$,
we have
\[
q_k(W)-q_0(W)
=\nabla h_k(X)+e_k(X)\{Y-\mu(X)\}-a_k(X)h_k(X).
\]
Thus the product-rate condition alone does not imply moment
stability. Under both conditions, we obtain asymptotic normality.

\begin{theorem}[Cross-fitted asymptotic normality]
\label{thm:crossfit}
Under Assumptions~\ref{ass:app-stein} and~\ref{ass:crossfit},
\begin{equation}
\sqrt n(\hat\theta-\theta_\PP)
=\frac1{\sqrt n}\sum_{i=1}^n\psi_0(W_i)+o_P(1)
\ \xrightarrow{d}\ N(0,V),
\qquad V=\E_P[\psi_0\psi_0^\top].
\label{eq:main-clt}
\end{equation}
Moreover, with $\hat\psi_i=q_{k(i)}(W_i)-\hat\theta$,
\begin{equation}
\hat V=\frac1n\sum_{i=1}^n\hat\psi_i\hat\psi_i^\top
\xrightarrow{P}V.
\label{eq:app-variance}
\end{equation}
\end{theorem}

\begin{proof}
Let $P_{n,k}$ denote the fold-$k$ empirical mean. Add and subtract $q_0$ to obtain
\[
\hat\theta-\theta_\PP
=(P_n-P)q_0
+\sum_{k=1}^K\frac{n_k}{n}(P_{n,k}-P)(q_k-q_0)
+\sum_{k=1}^K\frac{n_k}{n}P(q_k-q_0).
\]
Conditional on the observations outside fold $k$, its centered term has second moment at most $n_k^{-1}\|q_k-q_0\|_{L^2(P)}^2$. Conditional Chebyshev and fixed $K$ therefore make the middle sum $o_P(n^{-1/2})$. Proposition~\ref{prop:exact-bias} and the product condition give the same bound for the last sum. The multivariate central limit theorem gives~\eqref{eq:main-clt}.

For the covariance estimate, conditional Markov and moment stability imply
$n^{-1}\sum_i\|q_{k(i)}(W_i)-q_0(W_i)\|^2=o_P(1)$.
The law of large numbers gives $n^{-1}\sum_i\|q_0(W_i)\|^2=O_P(1)$, so Cauchy--Schwarz makes the difference between the fitted and true empirical second-moment matrices $o_P(1)$. Centering by the consistent $\hat\theta$, followed by the law of large numbers for $\psi_0\psi_0^\top$, proves~\eqref{eq:app-variance}.
\end{proof}

\paragraph{Normalization and local response.}
We estimate the average gradient $\theta_\PP$, but use its normalized
direction for steering. The following bound connects estimation error
in $\hat\theta$ to the loss in the steering objective from using the estimated direction
rather than the first-order optimal direction.

Suppose $\mu$ is twice continuously differentiable near all shifts
$X+u$ with $\|u\|\leq\delta_0$, $\E_\PP\|\nabla\mu(X)\|<\infty$, and
\[
B=\E_\PP\!\left[\sup_{\|u\|\leq\delta_0}
    \|\nabla^2\mu(X+u)\|_{\mathrm{op}}\right]<\infty.
\]
For nonzero $\theta_\PP$ and $\hat\theta$, let
$v^\star=\theta_\PP/\|\theta_\PP\|$ and
$\hat v=\hat\theta/\|\hat\theta\|$. Then, for $0\leq\delta\leq\delta_0$,
\begin{equation}
J(\delta,v^\star)-J(\delta,\hat v)
\leq \frac{2\delta\|\hat\theta-\theta_\PP\|^2}{\|\theta_\PP\|}
     +B\delta^2.
\label{eq:direction-error}
\end{equation}
The first term captures the cost of estimating the direction, which
is quadratic in gradient-estimation error for fixed $\theta_\PP\ne0$.
The second term accounts for curvature over a finite steering step.
Thus, for small steering steps, accurate gradient estimation limits the loss in the steering objective relative to $v^\star$.

For each unit vector $v$, Taylor's theorem gives
\[
\left|J(\delta,v)-J(0,v)-\delta\theta_\PP^\top v\right|
\leq \frac{B\delta^2}{2}.
\]
Combine these bounds with
\[
\theta_\PP^\top(v^\star-\hat v)
=\frac{\|\theta_\PP\|}{2}\|v^\star-\hat v\|^2
\leq \frac{2\|\hat\theta-\theta_\PP\|^2}{\|\theta_\PP\|}.
\]

\section{Worked examples and Gaussian special cases}
\label{app:worked-example}

We now consider several special cases to illustrate the key ideas.

\paragraph{Single-index outcome model.}
Let $X\sim N(0,\Sigma)$ with positive-definite $\Sigma$, and let $\mu(x)=g(\beta^\top x)$ satisfy the Stein conditions in Appendix~\ref{app:regularity}. Then
\[
\theta_\PP=c\beta,
\qquad c=\E_\PP[g'(\beta^\top X)],
\qquad \operatorname{Cov}_P(X,Y)=\Sigma\theta_\PP.
\]
Now consider a linear regression outcome model $m(x)=b^\top x+c_0$ and a working score $a(x)=Ax$, with $b,c_0,A$ fixed. Their population orthogonalized moment is
\begin{align}
\E_P[q(W;m,a)]
&=b+A\operatorname{Cov}_P(X,Y)-A\Sigma b \notag\\
&=\theta_\PP+(I-A\Sigma)(b-\theta_\PP),
\label{eq:gaussian-correction}
\end{align}
where the intercept drops out because $\E_\PP[X]=0$.
If $A=\Sigma^{-1}$, the bias vanishes for every slope $b$.
If $b=\theta_\PP$, it vanishes for every $A$, even when
$m\neq\mu$. Thus the population error factors into the
score error $I-A\Sigma$ and the slope discrepancy
$b-\theta_\PP$.

The population least-squares slope is
$b_{\mathrm{OLS}}
=\Sigma^{-1}\operatorname{Cov}_P(X,Y)=\theta_\PP$,
even for nonlinear $g$. With this slope and the true
Gaussian score, the outcome, score, and orthogonalized
population moments all equal $\theta_\PP$.

\paragraph{Binary contrast under a mixture distribution.}
Now let $Y=L\sim\operatorname{Bernoulli}(\pi)$, $0<\pi<1$, and suppose
\[
X\mid L=\ell\sim N(m_\ell,\Sigma),\qquad \ell\in\{0,1\}.
\]
The evaluation distribution is the mixture $\PP=(1-\pi)\PP_0+\pi\PP_1$. With $\Delta=m_1-m_0$ and $\beta=\Sigma^{-1}\Delta$, Bayes' rule gives
\[
\mu_L(x)=\Pr(L=1\mid X=x)=\operatorname{logit}^{-1}(c_0+\beta^\top x),
\qquad
\theta_\PP=\beta\,\E_\PP[\mu_L(X)\{1-\mu_L(X)\}].
\]
For $\Delta\ne0$, the average gradient therefore has the LDA direction. We can then obtain the (negative) score by differentiating: 
\begin{equation}
\alpha(x)=\Sigma^{-1}\{x-m_0-\mu_L(x)\Delta\}.
\label{eq:mixture-score}
\end{equation}

To compare this with a Gaussian working score, write
$s=\pi(1-\pi)$, $\kappa=\Delta^\top\Sigma^{-1}\Delta$, and
$\bar x=\E_{\PP}[X]$. The pooled covariance and population
least-squares slope satisfy
\[
\Sigma_X=\Sigma+s\Delta\Delta^\top,
\qquad
b_{\mathrm{OLS}}
=s\Sigma_X^{-1}\Delta
=\frac{s}{1+s\kappa}\,\beta.
\]
For the pooled Gaussian working score
$a_G(x)=\Sigma_X^{-1}(x-\bar x)$ and any affine regression $m$,
the slope terms in the orthogonalized moment cancel, giving
\[
\E_P[q(W;m,a_G)]=b_{\mathrm{OLS}}.
\]
For $\Delta\ne0$, this vector has the same LDA direction as
$\theta_{\PP}$, although their magnitudes generally differ.
Using the true mixture score instead recovers $\theta_{\PP}$
for every affine $m$.

\section{The average gradient outer product}
\label{sec:agop}

The first-order target $\theta_\PP$ averages gradients into
a single vector. An alternative is the \textit{average gradient
outer product}
\citep[AGOP; e.g.,][]{beaglehole2026universal}:
\begin{equation}
G_\PP=\E_\PP[\nabla\mu(X)\nabla\mu(X)^\top]
=\theta_\PP\theta_\PP^\top+\operatorname{Cov}_\PP[\nabla\mu(X)].
\label{eq:agop}
\end{equation}
For a unit vector $v$,
$v^\top G_\PP v=\E_\PP[(v^\top\nabla\mu(X))^2]$.
A leading eigenvector therefore maximizes average squared
first-order sensitivity. The second-order Taylor term in
the mean response, by contrast, depends on $\E_\PP[\nabla^2\mu(X)]$.

\paragraph{Orthogonalized estimation.} We can apply the same logic from the average gradient to estimating AGOP.
For twice-differentiable $m$ and a vector field $a$, define
\begin{equation}
M_{a,m}(x)=a(x)\nabla m(x)^\top+\nabla m(x)a(x)^\top-2\nabla^2m(x).
\label{eq:agop-M}
\end{equation}
The corresponding one-step moment and estimator are
\begin{equation}
Q_G(W;m,a)=\nabla m(X)\nabla m(X)^\top+M_{a,m}(X)\{Y-m(X)\},
\qquad
\hat G=\frac1n\sum_i Q_G(W_i;m_{k(i)},a_{k(i)}).
\label{eq:agop-one-step}
\end{equation}
As before, fitted functions are evaluated on held-out
observations from the same joint distribution $P$.
For this appendix, suppose that $\mu$ and the fitted
regressions belong to a fixed linear class
$\mathcal H_G\subseteq\mathcal H$ of twice continuously
differentiable functions with square-integrable gradients.
In addition to Assumption~\ref{ass:app-stein}, assume that the same
representer $\alpha$ satisfies
\[
\E_{\PP}[\nabla\{(m-\mu)\partial_jm\}]
=
\E_{\PP}[\alpha(m-\mu)\partial_jm],
\qquad
m\in\mathcal H_G,\quad j=1,\ldots,d,
\]
and that all displayed terms are integrable. We then have the following result. 
\begin{proposition}[Quadratic remainder]
\label{prop:agop-remainder}
Let $h=m-\mu$ and $e=a-\alpha$. Under these conditions,
\begin{equation}
\E_P[Q_G(W;m,a)]-G_\PP
=-\E_\PP[\nabla h\nabla h^\top]
-\E_\PP\!\left[h\{e\nabla m^\top+\nabla m e^\top\}\right].
\label{eq:agop-remainder}
\end{equation}
\end{proposition}

\begin{proof}
Applying the Stein identity to $h\partial_jm$, coordinate by coordinate, gives
\[
\E_\PP[hM_{\alpha,m}]
=\E_\PP[\nabla h\nabla m^\top+\nabla m\nabla h^\top].
\]
Since $\E_P[Y-m(X)\mid X]=-h(X)$, substituting this identity into~\eqref{eq:agop-one-step} and using $\nabla m=\nabla\mu+\nabla h$ leaves $G_\PP-\E_\PP[\nabla h\nabla h^\top]$. Replacing $\alpha$ by $a=\alpha+e$ contributes the second term in~\eqref{eq:agop-remainder}.
\end{proof}
Along admissible perturbations $m_t=\mu+tb$ and
$a_t=\alpha+tu$, with $b\in\mathcal H_G$, the remainder
is $O(t^2)$, establishing Neyman orthogonality.
The moment remains unbiased when $m=\mu$, but it is
not doubly robust, since it has bias
$-\E_\PP[\nabla h\nabla h^\top]$ even when $a=\alpha$.

The remainder also satisfies
\[
\bigl\|\E_P[Q_G(W;m,a)]-G_{\PP}\bigr\|_{\mathrm{op}}
\leq
\|\nabla h\|_2^2
+2\E_{\PP}\!\left[|h|\,\|e\|\,\|\nabla m\|\right].
\]
For estimated nuisances, making both terms
$o_P(n^{-1/2})$ controls the bias at the root-$n$ scale.
In particular, the first term is $o_P(n^{-1/2})$ when
$\|\nabla h\|_2=o_P(n^{-1/4})$.
A corresponding asymptotic normality result would also
require finite variance and cross-fitted $L^2(P)$ stability
of the matrix-valued moment.

\section{Additional experimental details}
\label{app:experiments}

This appendix gives the protocol for each experiment in Section~\ref{sec:experiments} and the uncertainty calculations. The supplementary material contains relevant scripts and outputs.

\subsection{Estimator as implemented}
\label{app:estimator}

For the outcome models we used CAA, linear ridge regression, and kernel ridge regression.
For the score models, we considered three approaches, each fitted in the span of the leading 64 principal components and lifted back with the orthonormal loadings. 
Specifically, we used the KDE score $-\nabla\log\hat p$ for a Gaussian kernel density estimate with median-heuristic bandwidth, the RFF score which is a Riesz regression on 512 random Fourier features with ridge penalty $10^{-2}$, and the isotropic score, which is the residual of the centered activation after regressing it on $\hat v^\top x$. 

Throughout the experiments we implement Equation~\eqref{eq:orth} with the outcome model and score above. 
The score scale is treated as a hyperparameter and chosen on validation, from $\{0.25,0.5,1,2, 4, 8\}$.
For the outcome models, hyperparameters are chosen by five-fold cross-validation grouped by statement on the training data split. Validation data select the steering layer, sign, and strength, the score model, and the score scale.

\subsection{Public opinion steering}

\paragraph{Data and splits.}
Questions come from OpinionQA~\citep{santurkar2023whose}, restricted to the 484 questions with Republican and Democratic answer distributions. Each seed shuffles the questions and splits them into 284 training, 100 validation, and 100 test questions. Seeds $0$--$4$ give five repetitions, and each seed's split is shared by all twelve model--target settings.

\paragraph{Prompts and response.}
Training prompts read ``Answer the following question as if in politics today, you considered yourself a \{party\}.'' followed by the question and ``Answer:''. Evaluation uses the question followed by ``Answer:''. The training response is the target group's expected zero-based option index, $\sum_k k\,p_k$, where $p_k$ is the group's share choosing the $k$th displayed option. For nominal options this proxy depends on the display order and does not identify the full answer distribution. The model's answer distribution sums the probability of each option letter, with and without a leading space, and renormalizes over the offered options.

\paragraph{Selection and evaluation.}
For each estimator, validation TV selects the layer from $\{6,8,\ldots,24\}$, a signed strength multiplier from $\{0.25,0.5,1,\ldots,256\}$, the score model, and the score scale. Validation KL makes a separate selection. Strength is the multiplier times the standard deviation of the training projections onto the direction. The steering vector is added to the last-token residual. TV is half the $L_1$ distance; KL is $\mathrm{KL}(\text{target}\,\|\,\text{model})$. Each test metric averages the 100 test questions equally.

Table~\ref{tab:opinion-outcomes} repeats the comparison of Table~\ref{tab:opinion} with linear ridge, KRR, and RFM as the outcome model. Each is fit to the same training response as CAA, and the Neyman estimator built on it chooses the score model and the size of the correction on validation, including no correction. These runs reproduce the CAA plug-in of Table~\ref{tab:opinion} exactly.

\begin{table*}[t]
  \centering
  \small
  \setlength{\tabcolsep}{4.5pt}
  \caption{OpinionQA with linear ridge, KRR and RFM outcome models: plug-in (``Plug'') and Neyman (``Ney.'') estimators built on each, test TV averaged over both targets and five seeds, then all-model TV and KL; lower is better. ``Gain'' is the mean of the setting-level relative reductions, as in Table~\ref{tab:opinion}. All-model 95\% intervals, resampling seeds: linear $[5.4, 7.0]$ (TV) and $[12.8, 18.2]$ (KL); KRR $[4.7, 7.8]$ and $[11.5, 17.4]$; RFM $[4.9, 8.1]$ and $[10.6, 17.7]$.}
  \label{tab:opinion-outcomes}
  \begin{tabular}{lccr|ccr|ccr}
    \toprule
    & \multicolumn{3}{c|}{Linear} & \multicolumn{3}{c|}{KRR} & \multicolumn{3}{c}{RFM} \\
    \cmidrule(lr){2-4}\cmidrule(lr){5-7}\cmidrule(lr){8-10}
    Model & Plug & Ney. & Gain & Plug & Ney. & Gain & Plug & Ney. & Gain \\
    \midrule
    Llama-3.1-8B & .268 & \textbf{.260} & 2.8 & .268 & \textbf{.259} & 3.1 & .268 & \textbf{.260} & 3.0 \\
    Qwen3-4B-Base & .257 & \textbf{.255} & 0.2 & .256 & \textbf{.252} & 1.6 & .256 & \textbf{.252} & 1.7 \\
    Qwen3-8B & .331 & \textbf{.281} & 14.6 & .338 & \textbf{.276} & 18.2 & .338 & \textbf{.275} & 18.5 \\
    Qwen3-8B-Base & .246 & \textbf{.237} & 3.6 & .247 & \textbf{.235} & 4.7 & .246 & \textbf{.235} & 4.3 \\
    Qwen3.5-9B & .295 & \textbf{.261} & 11.3 & .291 & \textbf{.269} & 6.9 & .291 & \textbf{.269} & 7.2 \\
    Qwen3.5-9B-Base & .269 & \textbf{.257} & 4.5 & .268 & \textbf{.261} & 2.5 & .268 & \textbf{.257} & 4.0 \\
    \midrule
    All TV & .278 & \textbf{.259} & 6.2 & .278 & \textbf{.259} & 6.2 & .278 & \textbf{.258} & 6.4 \\
    All KL & .286 & \textbf{.232} & 15.5 & .288 & \textbf{.235} & 14.6 & .289 & \textbf{.238} & 14.3 \\
    \bottomrule
  \end{tabular}
\end{table*}

\subsection{Truthfulness steering}

\paragraph{Data.}
The nine datasets are \textsc{Cities}, \textsc{Sp-En Trans.}, \textsc{Neg Cities}, \textsc{Neg Sp-En Trans.}, \textsc{Cities Conj.}, \textsc{Cities Disj.}, \textsc{Companies}, \textsc{Common Claim}, and \textsc{Counterfact}~\citep{marks2024geometry,burns2023discovering}. Each is split 60/20/20 by a seeded permutation and capped at 400/150/300 statements. Seeds 0 and 1 vary both the split and the subsampling.

\paragraph{Prompts and response orientation.}
The three variants of statement $s$ are the neutral prompt ``$s$ This statement is:'', the same statement preceded by ``Answer this truthfully with TRUE or FALSE.'', and the statement preceded by an instruction to give the deliberately wrong answer. 

\paragraph{Selection and metrics.}
Validation selects the layer from $\{12,16,20,24\}$, a signed multiplier from $\{4,8,16,32,64,128\}$, the score model, and the score scale, with no steering among the candidates. The TV and KL columns use the configurations that minimize validation TV and validation KL, respectively. Reported accuracy is for the TV-selected configuration.

\begin{table}[ht]
  \centering
  \small
  \caption{Truthfulness by dataset, mean test error over six models and two seeds; lower is better.}
  \label{tab:truth-datasets}
  \begin{tabular}{lcccc}
    \toprule
    & \multicolumn{2}{c}{TV} & \multicolumn{2}{c}{KL} \\
    \cmidrule(lr){2-3}\cmidrule(lr){4-5}
    Dataset & CAA & Neyman & CAA & Neyman \\
    \midrule
    \textsc{Cities} & 0.221 & 0.121 & 0.295 & 0.154 \\
    \textsc{Sp-En Trans.} & 0.350 & 0.264 & 0.550 & 0.346 \\
    \textsc{Neg Cities} & 0.365 & 0.297 & 0.506 & 0.377 \\
    \textsc{Neg Sp-En Trans.} & 0.406 & 0.370 & 0.591 & 0.510 \\
    \textsc{Cities Conj.} & 0.280 & 0.195 & 0.379 & 0.254 \\
    \textsc{Cities Disj.} & 0.399 & 0.351 & 0.554 & 0.476 \\
    \textsc{Companies} & 0.381 & 0.334 & 0.654 & 0.496 \\
    \textsc{Common Claim} & 0.402 & 0.378 & 0.675 & 0.573 \\
    \textsc{Counterfact} & 0.403 & 0.370 & 0.660 & 0.585 \\
    \bottomrule
  \end{tabular}
\end{table}

\subsection{AxBench}

\paragraph{Setup.}
We use Concept500 with Gemma-2-2B-it and AxBench's released examples: per concept, 72 positive examples drawn from the concept's genre and 216 negatives split equally across text, code, and math. Activations are mean layer-ten residuals over response tokens. Directions are unit-normalized and added to the layer-ten residual at every position, with intervention $x+s\,\hat\theta$ for strength $s$. The layer is fixed at ten, following AxBench. Generation samples 128 new tokens at temperature $1$ with a fixed seed per row.

\paragraph{Judge.}
We use AxBench's three judge prompts, its rating parser, and its rule that the steering score is the harmonic mean of the three components, or zero if any component is zero. Qwen3-8B serves as judge, with greedy decoding and its thinking mode off.

\paragraph{Validation and test instructions.}
Each concept has ten AlpacaEval test instructions, as in AxBench, and ten disjoint AlpacaEval validation instructions. The strength, from $\{24,32,48,64,80,96,112\}$, and the choice among CAA and the two Neyman directions are selected on validation only, separately within each genre. Validation selects CAA at strength $96$ for text, CAA at $64$ for code, and CAA at $80$ for math. For the Neyman estimator it selects $\hat v_g$ at strength $80$ for text and CAA for code and math.

\paragraph{Prompting baseline.}
The prompting baseline uses AxBench's prompting template, which prefixes the instruction with a request to answer with content related to the concept. It uses the same test instructions and generation settings as the steering runs. Its components are concept $0.468$, instruction $1.845$, and fluency $1.181$.

\subsection{Compute}
All runs used one node with four NVIDIA H100 80GB GPUs, with one job per GPU. AxBench judging used vLLM.

\end{document}